\documentclass[11pt]{article}
\usepackage{coling2016}
\usepackage{times}
\usepackage{url}
\usepackage{latexsym}

\usepackage{graphicx}
\usepackage{amsmath}
\usepackage{booktabs}
\usepackage{tabularx}

\title{Attending to Characters in Neural Sequence Labeling Models}

\author{\hspace{-1.5cm}Marek Rei\\
        \hspace{-1.5cm}The ALTA Institute\\
        \hspace{-1.5cm}Computer Laboratory\\
        \hspace{-1.5cm}University of Cambridge\\
        \hspace{-1.5cm}United Kingdom\\
        \hspace{-1.5cm}{\tt marek.rei@cl.cam.ac.uk}
\And
        \hspace{-1.5cm}Gamal K.O. Crichton \hspace{1cm} Sampo Pyysalo\\
        \hspace{-1.5cm}Language Technology Lab\\
        \hspace{-1.5cm}Dept. of Theoretical \& Applied Linguistics\\
        \hspace{-1.5cm}University of Cambridge\\
        \hspace{-1.5cm}United Kingdom\\
        \hspace{-1.5cm}{\tt \{gkoc2,smp66\}@cam.ac.uk} \\}

\date{}

\begin{document}
\maketitle
\begin{abstract}

Sequence labeling architectures use word embeddings for capturing similarity, but suffer when handling previously unseen or rare words.
We investigate character-level extensions to such models and propose a novel architecture for combining alternative word representations.
By using an attention mechanism, the model is able to dynamically decide how much information to use from a word- or character-level component. 
We evaluated different architectures on a range of sequence labeling datasets, and character-level extensions were found to improve performance on every benchmark.
In addition, the proposed attention-based architecture delivered the best results even with a smaller number of trainable parameters.


\end{abstract}

\section{Introduction}

\blfootnote{
     \hspace{-0.65cm}  
     This work is licenced under a Creative Commons 
     Attribution 4.0 International Licence.\\
     Licence details:
     \url{http://creativecommons.org/licenses/by/4.0/}
}

Many NLP tasks, including named entity recognition (NER), part-of-speech (POS) tagging and shallow parsing can be framed as types of sequence labeling. 
The development of accurate and efficient sequence labeling models is thereby useful for a wide range of downstream applications.
Work in this area has traditionally involved task-specific feature engineering -- for example, integrating gazetteers for named entity recognition, or using features from a morphological analyser in POS-tagging.
Recent developments in neural architectures and representation learning have opened the door to models that can discover useful features automatically from the data.
Such sequence labeling systems are applicable to many tasks, using only the surface text as input, yet are able to achieve competitive results \cite{Collobert2011,Irsoy2014a}.

Current neural models generally make use of word embeddings, which allow them to learn similar representations for semantically or functionally similar words. While this is an important improvement over count-based models, they still have weaknesses that should be addressed. 
The most obvious problem arises when dealing with out-of-vocabulary (OOV) words -- if a token has never been seen before, then it does not have an embedding and the model needs to back-off to a generic OOV representation. Words that have been seen very infrequently have embeddings, but they will likely have low quality due to lack of training data. 
The approach can also be sub-optimal in terms of parameter usage -- for example, certain suffixes indicate more likely POS tags for these words, but this information gets encoded into each individual embedding as opposed to being shared between the whole vocabulary.

In this paper, we construct a task-independent neural network architecture for sequence labeling, and then extend it with two different approaches for integrating character-level information. By operating on individual characters, the model is able to infer representations for previously unseen words and share information about morpheme-level regularities. We propose a novel architecture for combining character-level representations with word embeddings using a gating mechanism, also referred to as \textit{attention}, which allows the model to dynamically decide which source of information to use for each word. In addition, we describe a new objective for model training where the character-level representations are optimised to mimic the current state of word embeddings.

We evaluate the neural models on 8 datasets from the fields of NER, POS-tagging, chunking and error detection in learner texts. Our experiments show that including a character-based component in the sequence labeling model provides substantial performance improvements on all the benchmarks. In addition, the attention-based architecture achieves the best results on all evaluations, while requiring a smaller number of parameters.





\section{Bidirectional LSTM for sequence labeling}
\label{sec:baseline}

We first describe a basic word-level neural network for sequence labeling, following the models described by \newcite{Lample2016} and \newcite{Rei2016}, and then propose two alternative methods for incorporating character-level information.

\begin{figure}[t]
    \centering
	\includegraphics[width=0.5\linewidth]{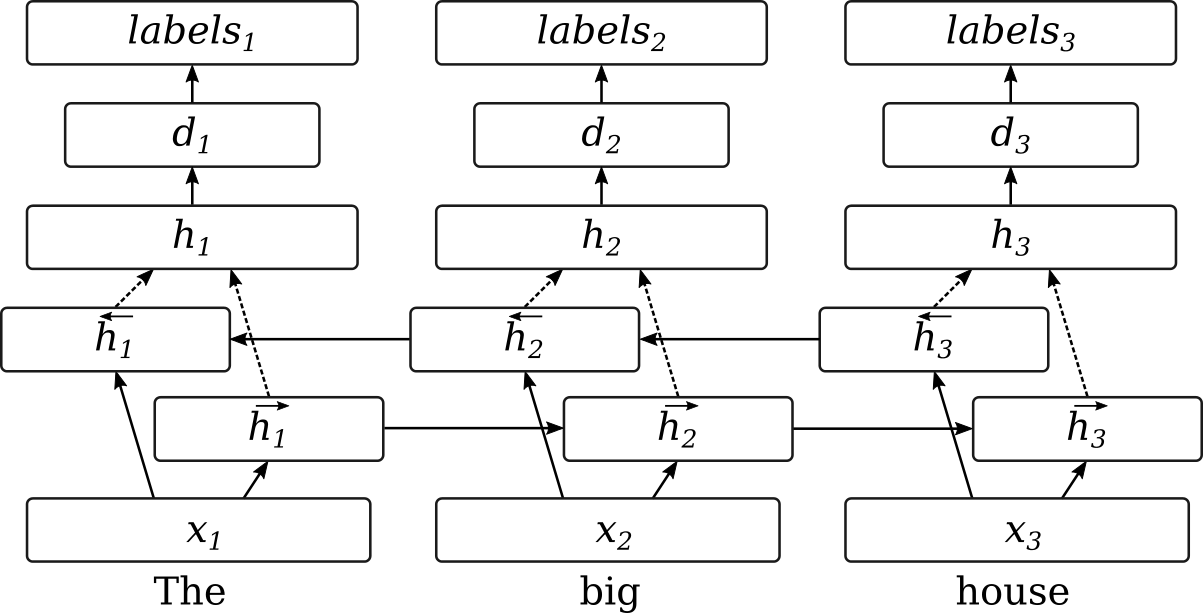}
	\caption{Neural sequence labeling model. Word embeddings are given as input; a bidirectional LSTM produces context-dependent representations; the information is passed through a hidden layer and the output layer. The outputs are either probability distributions for softmax, or confidence scores for CRF. }
	\label{fig:baseline}
\end{figure}

Figure \ref{fig:baseline} shows the general architecture of the sequence labeling network. The model receives a sequence of tokens $(w_1, ..., w_T)$ as input, and predicts a label corresponding to each of the input tokens.
The tokens are first mapped to a distributed vector space, resulting in a sequence of word embeddings $(x_1, ..., x_T)$.
Next, the embeddings are given as input to two LSTM \cite{Hochreiter1997} components moving in opposite directions through the text, creating context-specific representations. The respective forward- and backward-conditioned representations are concatenated for each word position, resulting in representations that are conditioned on the whole sequence:

\begin{equation}
\overrightarrow{h_t} = LSTM(x_t, \overrightarrow{h_{t-1}})\hspace{3em}
\overleftarrow{h_t} = LSTM(x_t, \overleftarrow{h_{t+1}})\hspace{3em}
h_t = [\overrightarrow{h_t};\overleftarrow{h_t}]
\end{equation}

We include an extra narrow hidden layer on top of the LSTM, which proved to be a useful modification based on development experiments. An additional hidden layer allows the model to detect higher-level feature combinations, while constraining it to be small forces it to focus on more generalisable patterns:

\begin{equation}
d_t = tanh(W_d h_t)
\end{equation}

\noindent where $W_d$ is a weight matrix between the layers, and the size of $d_t$ is intentionally kept small.

Finally, to produce label predictions, we use either a softmax layer or a conditional random field (CRF, \newcite{Lafferty2001}). The softmax calculates a normalised probability distribution over all the possible labels for each word:

\begin{equation}
P(y_t = k | d_t) = \frac{e^{W_{o,k} d_t}}{\sum_{\tilde{k} \in K} e^{W_{o,\tilde{k}} d_t}}
\end{equation}

\noindent where $P (y_t = k|d_t )$ is the probability of the label of the $t$-th word ($y_t$) being $k$, $K$ is the set of all possible labels, and $W_{o,k}$ is the $k$-th row of output weight matrix $W_o$. To optimise this model, we minimise categorical crossentropy, which is equivalent to minimising the negative log-probability of the correct labels:

\begin{equation}
E = - \sum_{t=1}^{T} log(P(y_t| d_t))
\end{equation}

Following \newcite{Huang2015}, we can also use a CRF as the output layer, which conditions each prediction on the previously predicted label. In this architecture, the last hidden layer is used to predict confidence scores for the word having each of the possible labels. A separate weight matrix is used to learn transition probabilities between different labels, and the Viterbi algorithm is used to find an optimal sequence of weights. Given that $y$ is a sequence of labels $[y_1, ..., y_T]$, then the CRF score for this sequence can be calculated as:

\begin{equation}
s(y) = \sum_{t=1}^T A_{t,y_t} + \sum_{t=0}^T B_{y_t,y_{t+1}}
\end{equation}

\begin{equation}
A_{t,y_t} = W_{o,y_t} d_t
\end{equation}

\noindent where $A_{t,y_t}$ shows how confident the network is that the label on the $t$-th word is $y_t$. $B_{y_t,y_{t+1}}$ shows the likelihood of transitioning from label $y_t$ to label $y_{t+1}$, and these values are optimised during training. The output from the model is the sequence of labels with the largest score $s(y)$, which can be found efficiently using the Viterbi algorithm. In order to optimise the CRF model, the loss function maximises the score for the correct label sequence, while minimising the scores for all other sequences:

\begin{equation}
E = - s(y) + log \sum_{\tilde{y} \in \widetilde{Y}} e^{s(\tilde{y})}
\end{equation}

\noindent where $\widetilde{Y}$ is the set of all possible label sequences.

\section{Character-level sequence labeling}
\label{sec:concat}

Distributed embeddings map words into a space where semantically similar words have similar vector representations, allowing the models to generalise better.
However, they still treat words as atomic units and ignore any surface- or morphological similarities between different words. By constructing models that operate over individual characters in each word, we can take advantage of these regularities. This can be particularly useful for handling unseen words -- for example, if we have never seen the word \textit{cabinets} before, a character-level model could still infer a representation for this word if it has previously seen the word \textit{cabinet} and other words with the suffix \textit{-s}. In contrast, a word-level model can only represent this word with a generic out-of-vocabulary representation, which is shared between all other unseen words.

\begin{figure}[t]
    \centering
	\includegraphics[width=0.85\linewidth]{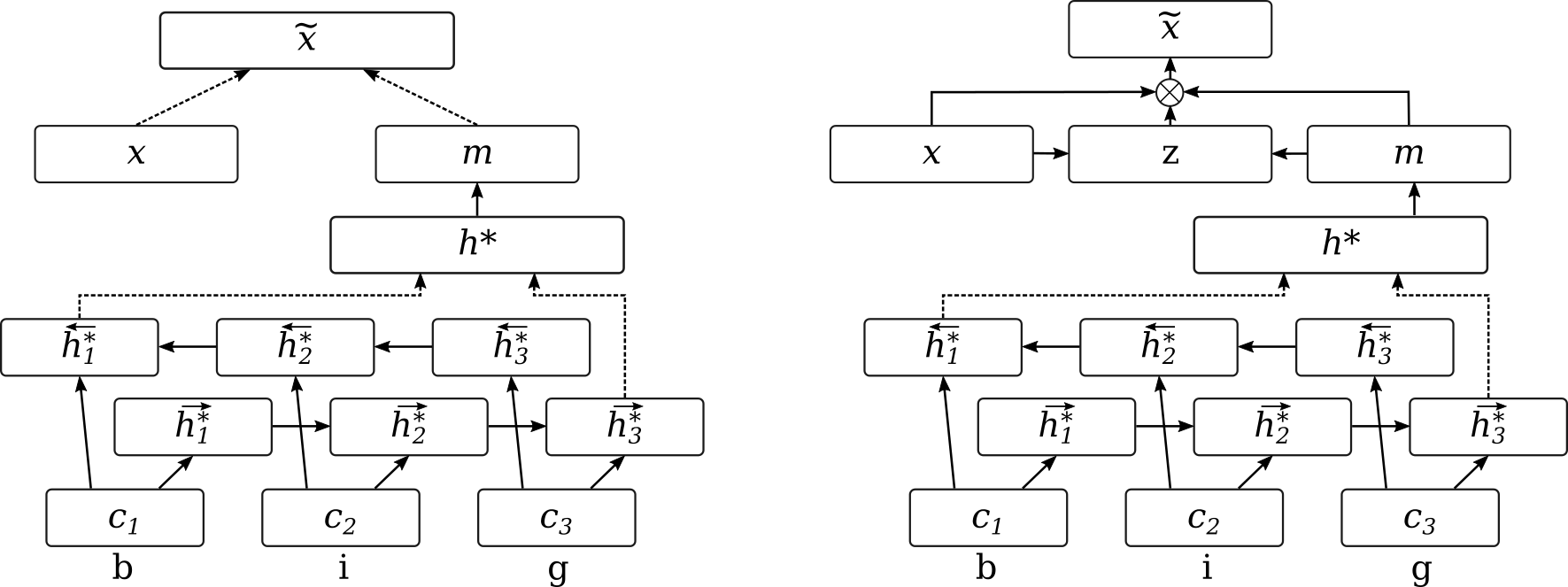}
	\caption{Left: concatenation-based character architecture. Right: attention-based character architecture. The dotted lines indicate vector concatenation.}
	\label{fig:char}
\end{figure}

Research into character-level models is still in fairly early stages, and models that operate exclusively on characters are not yet competitive to word-level models on most tasks. However, instead of fully replacing word embeddings, we are interested in combining the two approaches, thereby allowing the model to take advantage of information at both granularity levels.
The general outline of our approach is shown in Figure \ref{fig:char}. Each word is broken down into individual characters, these are then mapped to a sequence of character embeddings $(c_1, ..., c_R)$, which are passed through a bidirectional LSTM:

\begin{equation}
\overrightarrow{h^*_i} = LSTM(c_i, \overrightarrow{h^*_{i-1}}) \hspace{3em}
\overleftarrow{h^*_i} = LSTM(c_i, \overleftarrow{h^*_{i+1}})
\end{equation}

We then use the last hidden vectors from each of the LSTM components, concatenate them together, and pass the result through a separate non-linear layer.

\begin{equation}
h^* = [\overrightarrow{h^*_R};\overleftarrow{h^*_1}] \hspace{3em}
m = tanh(W_m h^*)
\end{equation}

\noindent where $W_m$ is a weight matrix mapping the concatenated hidden vectors from both LSTMs into a joint word representation $m$, built from individual characters.

We now have two alternative feature representations for each word -- $x_t$ from Section \ref{sec:baseline} is an embedding learned on the word level, and $m^{(t)}$ is a representation dynamically built from individual characters in the $t$-th word of the input text. Following \newcite{Lample2016}, one possible approach is to concatenate the two vectors and use this as the new word-level representation for the sequence labeling model:

\begin{equation}
\widetilde{x} = [x; m]
\end{equation}

\noindent This approach, also illustrated in Figure \ref{fig:char}, assumes that the word-level and character-level components learn somewhat disjoint information, and it is beneficial to give them separately as input to the sequence labeler. 

\section{Attention over character features}
\label{sec:attn}

Alternatively, we can have the word embedding and the character-level component learn the same semantic features for each word. Instead of concatenating them as alternative feature sets, we specifically construct the network so that they would learn the same representations, and then allow the model to decide how to combine the information for each specific word.

We first construct the word representation from characters using the same architecture -- a bidirectional LSTM operates over characters, and the last hidden states are used to create vector $m$ for the input word. Instead of concatenating this with the word embedding, the two vectors are added together using a weighted sum, where the weights are predicted by a two-layer network:

\begin{equation}
z = \sigma(W^{(3)}_z tanh(W^{(1)}_{z} x + W^{(2)}_{z} m)) \hspace{3em}
\widetilde{x} = z\cdot x + (1-z) \cdot m
\end{equation}

\noindent where $W^{(1)}_{z}$, $W^{(2)}_{z}$ and $W^{(3)}_{z}$ are weight matrices for calculating $z$, and $\sigma()$ is the logistic function with values in the range $[0,1]$. The vector $z$ has the same dimensions as $x$ or $m$, acting as the weight between the two vectors. It allows the model to dynamically decide how much information to use from the character-level component or from the word embedding. This decision is done for each feature separately, which adds extra flexiblity -- for example, words with regular suffixes can share some character-level features, whereas irregular words can store exceptions into word embeddings.
Furthermore, previously unknown words are able to use character-level regularities whenever possible, and are still able to revert to using the generic OOV token when necessary.

The main benefits of character-level modeling are expected to come from improved handling of rare and unseen words, whereas frequent words are likely able to learn high-quality word-level embeddings directly. We would like to take advantage of this, and train the character component to predict these word embeddings.
Our attention-based architecture requires the learned features in both word representations to align, and we can add in an extra constraint to encourage this. During training, we add a term to the loss function that optimises the vector $m$ to be similar to the word embedding $x$:

\begin{equation}
\label{eq:attncost}
\widetilde{E} = E + \sum_{t=1}^{T} g_t (1 - cos(m^{(t)}, x_t)) \hspace{3em}
g_t =
\begin{cases}
      0, & \text{if}\ w_t = OOV \\
      1, & \text{otherwise}
    \end{cases}
\end{equation}

\noindent Equation \ref{eq:attncost} maximises the cosine similarity between $m^{(t)}$ and $x_t$. Importantly, this is done only for words that are not out-of-vocabulary -- we want the character-level component to learn from the word embeddings, but this should exclude the OOV embedding, as it is shared between many words. We use $g_t$ to set this cost component to $0$ for any OOV tokens.

While the character component learns general regularities that are shared between all the words, individual word embeddings provide a way for the model to store word-specific information and any exceptions.
Therefore, while we want the character-based model to shift towards predicting high-quality word embeddings, it is not desireable to optimise the word embeddings towards the character-level representations. 
This can be achieved by making sure that the optimisation is performed only in one direction; in Theano \cite{Bergstra2010}, the \textit{disconnected\_grad} function gives the desired effect.

\section{Datasets}

\begin{table}
\centering
\begin{tabular}{llrrrr} \toprule
Name     & Task            & \# labels & \# train tokens & \# dev tokens & \# test tokens\\ \midrule
CoNLL00 & Chunking & 22 & 158,795 & 52,932 & 47,377 \\
CoNLL03 & NER & 8 & 203,621 & 51,362 & 46,435 \\
PTB-POS & POS &  48 & 912,344 & 131,768 & 129,654\\
FCEPUBLIC & Error det & 2 & 452,833 & 34,599 & 41,477 \\
BC2GM    & NER     & 3 & 355,405 & 71,042 & 143,465\\
CHEMDNER & NER & 3 & 891,948 & 886,324 & 766,033\\
JNLPBA   & NER & 11 & 445,090 & 47,461 & 101,039\\
GENIA-POS    & POS             & 42 & 397,690 & 52,697 & 50,556\\ \bottomrule
\end{tabular}
\caption{Details for each of the evaluation datasets.}
\label{tbl:datasets}
\end{table}

We evaluate the sequence labeling models and character architectures on 8 different datasets. Table \ref{tbl:datasets} contains information about the number of labels and dataset sizes for each of them.

\begin{itemize}
\item \textbf{CoNLL00}: The CoNLL-2000 dataset \cite{TjongKimSang2000} is a frequently used benchmark for the task of chunking. Wall Street Journal Sections 15-18 from the Penn Treebank are used for training, and Section 20 as the test data. As there is no official development set, we separated some of the training set for this purpose.

\item \textbf{CoNLL03}: The CoNLL-2003 corpus \cite{TjongKimSang2003} was created for the shared task on language-independent NER. 
We use the English section of the dataset, containing news stories from the Reuters Corpus\footnote{http://about.reuters.com/researchandstandards/corpus/}.

\item \textbf{PTB-POS}: The Penn Treebank POS-tag corpus \cite{Marcus1993b} contains texts from the Wall Street Journal, annotated for part-of-speech tags. The PTB label set includes 36 main tags and an additional 12 tags covering items such as punctuation.

\item \textbf{FCEPUBLIC}:
The publicly released subset of the First Certificate in English (FCE) dataset contains short essays written by language learners and manual corrections by examiners \cite{Yannakoudakis2011}. We use a version of this corpus converted into a binary error detection task, where each token is labeled as being correct or incorrect in the given context. 

\item \textbf{BC2GM}:
The BioCreative~II Gene Mention corpus \cite{Smith2008} consists of 20,000
sentences from biomedical publication abstracts and is annotated for
mentions of the names of genes, proteins and related entities
using a single NE class.

\item \textbf{CHEMDNER}:
The BioCreative~IV Chemical and Drug \cite{Krallinger2015} NER corpus consists of 10,000 abstracts annotated for mentions of
chemical and drug names using a single class.
We make use of the official splits provided by the shared task organizers.

\item \textbf{JNLPBA}:
The JNLPBA corpus \cite{Kim2004} consists of 2,404 biomedical abstracts and is annotated for mentions of five entity types:
\textsc{cell line}, \textsc{cell type}, \textsc{dna}, \textsc{rna},
and \textsc{protein}.
The corpus was derived from GENIA corpus entity annotations for use in
the shared task organized in conjuction with the BioNLP~2004 workshop.

\item \textbf{GENIA-POS}:
The GENIA corpus \cite{Ohta2002} is one of the most widely used resources for
biomedical NLP and has a rich set of annotations including parts of
speech, phrase structure syntax, entity mentions, and events. Here, we make use of the GENIA POS annotations,
which cover 2,000 PubMed abstracts (approx.\ 20,000 sentences).
We use the same 210-document test set as \newcite{Tsuruoka2005}, and additionally split off a sample of 210 from the remaining
documents as a development set.

\end{itemize}

\begin{table*}
\setlength\tabcolsep{10.5pt}
\begin{tabular}{r|rr|rr|rr|rr} \toprule
 & \multicolumn{2}{c|}{CoNLL00} & \multicolumn{2}{c|}{CoNLL03}  & \multicolumn{2}{c|}{PTB-POS}  & \multicolumn{2}{c}{FCEPUBLIC} \\ 
 & {\small DEV} & {\small TEST} & {\small DEV} & {\small TEST} & {\small DEV} & {\small TEST} & {\small DEV} & {\small TEST} \\ \midrule
Word-based & 91.48 & 91.23 & 86.89 & 79.86 & 96.29 & 96.42 & 46.58 & 41.24 \\
Char concat & 92.57 & 92.35 & 89.81 & 83.37 & 97.20 & 97.22 & 46.44 & 41.27 \\
Char attention & \textbf{92.92} & \textbf{92.67} & \textbf{89.91} & \textbf{84.09} & \textbf{97.22} & \textbf{97.27} & \textbf{47.17} & \textbf{41.88} \\ \bottomrule
\end{tabular}

\vspace{0.7cm}

\begin{tabular}{r|rr|rr|rr|rr} \toprule
 & \multicolumn{2}{c|}{BC2GM} & \multicolumn{2}{c|}{CHEMDNER} & \multicolumn{2}{c|}{JNLPBA} & \multicolumn{2}{c}{GENIA-POS} \\ 
 & {\small DEV} & {\small TEST} & {\small DEV} & {\small TEST} & {\small DEV} & {\small TEST} & {\small DEV} & {\small TEST} \\ \midrule
Word-based & 84.07 & 84.21 & 78.63 & 79.74 & 75.46 & 70.75 & 97.55 & 97.39 \\
Char concat & 87.54 & 87.75 & 82.80 & 83.56 & 76.82 & 72.24 & 98.59 & 98.49 \\
Char attention & \textbf{87.98} & \textbf{87.99} & \textbf{83.75} & \textbf{84.53} & \textbf{77.38} & \textbf{72.70} & \textbf{98.67} & \textbf{98.60} \\ \bottomrule
\end{tabular}
\caption{Comparison of word-based and character-based sequence labeling architectures on 8 datasets. The evaluation measure used for each dataset is specified in Section \ref{sec:evaluation}.}
\label{tab:results}
\end{table*}

\section{Experiment settings}
\label{sec:evaluation}

For data prepocessing, all digits were replaced with the character '0'. Any words that occurred only once in the training data were replaced by the generic OOV token for word embeddings, but were still used in the character-level components. 
The word embeddings were initialised with publicly available pretrained vectors, created using word2vec \cite{Mikolov2013a}, and then fine-tuned during model training. For the general-domain datasets we used 300-dimensional vectors trained on Google News\footnote{https://code.google.com/archive/p/word2vec/}; for the biomedical datasets we used 200-dimensional vectors trained on PubMed and PMC\footnote{http://bio.nlplab.org/}. The embeddings for characters were set to length $50$ and initialised randomly.

The LSTM layer size was set to $200$ in each direction for both word- and character-level components.
The hidden layer $d$ has size $50$, and the combined representation $m$ has the same length as the word embeddings.
CRF was used as the output layer for all the experiments -- we found that this gave most benefits to tasks with larger numbers of possible labels.
Parameters were optimised using AdaDelta \cite{Zeiler2012} with default learning rate $1.0$ and sentences were grouped into batches of size $64$. 
Performance on the development set was measured at every epoch and training was stopped if performance had not improved for 7 epochs; the best-performing model on the development set was then used for evaluation on the test set.
In order to avoid any outlier results due to randomness in the model initialisation, we trained each configuration with 10 different random seeds and present here the averaged results.

When evaluating on each dataset, we report the measures established in previous work.
Token-level accuracy is used for PTB-POS and GENIA-POS; $F_{0.5}$ score over the erroneous words for FCEPUBLIC; the official evaluation script for BC2GM which allows for alternative correct entity spans; and microaveraged mention-level $F_{1}$ score for the remaining datasets.

\section{Results}
\label{sec:results}

While optimising the hyperparameters for each dataset separately would likely improve individual performance, we conduct more controlled experiments on a task-independent model. Therefore, we use the same hyperparameters from Section \ref{sec:evaluation} on all datasets, and the development set is only used for the stopping condition.
With these experiments, we wish to determine 1) on which sequence labeling tasks do character-based models offer an advantange, and 2) which character-based architecture performs better.

Results for the different model architectures on all 8 datasets are shown in Table \ref{tab:results}. As can be seen, including a character-based component in the sequence labeling architecture improves performance on every benchmark. The NER datasets have the largest absolute improvement -- the model is able to learn character-level patterns for names, and also improve the handling of any previously unseen tokens. 

\begin{figure}[t]
    \centering
	\includegraphics[width=0.7\linewidth]{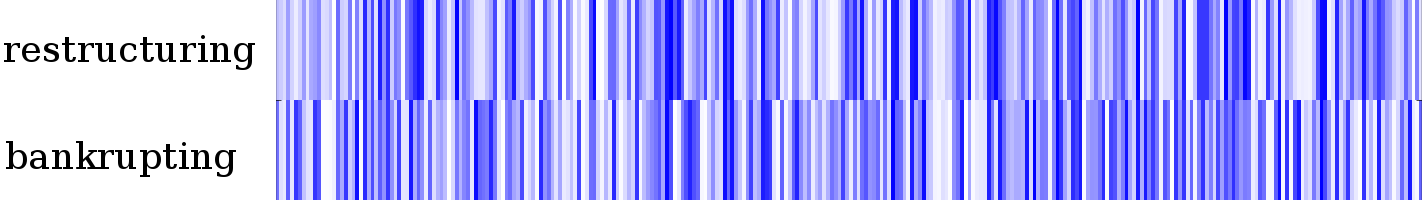}
	\caption{Visualisation of attention values for two words, trained on the PTB-POS dataset. Darker blue indicates features with higher weights for the character-level representation. \textit{Restructuring} was present in the vocabulary, while \textit{bankrupting} is an OOV.}
	\label{fig:attention}
\end{figure}

Compared to concatenating the word- and character-level representations, the attention-based character model outperforms the former on all evaluations. The mechanism for dynamically deciding how much character-level information to use allows the model to better handle individual word representations, giving it an advantage in the experiments. Visualisation of the attention values in Figure \ref{fig:attention} shows that the model is actively using character-based features, and the attention areas vary between different words.

The results of this general tagging architecture are competitive, even when compared to previous work using hand-crafted features. The network achieves 97.27\% on PTB-POS compared to 97.55\% by \newcite{Huang2015}, and 72.70\% on JNLPBA compared to 72.55\% by \newcite{Zhou2004}.
In some cases, we are also able to beat the previous best results -- 87.99\% on BC2GM compared to 87.48\% by \newcite{Campos2015}, and 41.88\% on FCEPUBLIC compared to 41.1\% by \newcite{Rei2016}.
\newcite{Lample2016} report a considerably higher result of 90.94\% on CoNLL03, indicating that the chosen hyperparameters for the baseline system are suboptimal for this specific task. Compared to the experiments presented here, their model used the IOBES tagging scheme instead of the original IOB, and embeddings pretrained with a more specialised method that accounts for word order.

It is important to also compare the parameter counts of alternative neural architectures, as this shows their learning capacity and indicates their time requirements in practice. Table \ref{tab:paramsizes} contains the parameter counts on three representative datasets. While keeping the model hyperparameters constant, the character-level models require additional parameters for the character composition and character embeddings. However, the attention-based model uses fewer parameters compared to the concatenation approach. When the two representations are concatenated, the overall word representation size is increased, which in turn increases the number of parameters required for the word-level bidirectional LSTM. Therefore, the attention-based character architecture achieves improved results even with a smaller parameter footprint.

\section{Related work}

There is a wide range of previous work on constructing and optimising neural architectures applicable to sequence labeling.
\newcite{Collobert2011} described one of the first task-independent neural tagging models using convolutional neural networks. They were able to achieve good results on POS tagging, chunking, NER and semantic role labeling, without relying on hand-engineered features. 
\newcite{Irsoy2014a} experimented with multi-layer bidirectional Elman-style recurrent networks, and found that the deep models outperformed conditional random fields on the task of opinion mining.
\newcite{Huang2015} described a bidirectional LSTM model with a CRF layer, which included hand-crafted features specialised for the task of named entity recognition.
\newcite{Rei2016} evaluated a range of neural architectures, including convolutional and recurrent networks, on the task of error detection in learner writing.
The word-level sequence labeling model described in this paper follows the previous work, combining useful design choices from each of them.
In addition, we extended the model with two alternative character-level architectures, and evaluated its performance on 8 different datasets.

Character-level models have the potential of capturing morpheme patterns, thereby improving generalisation on both frequent and unseen words. In recent years, there has been an increase in research into these models, resulting in several interesting applications.
\newcite{Ling2015b} described a character-level neural model for machine translation, performing both encoding and decoding on individual characters.
\newcite{Kim2016} implemented a language model where encoding is performed by a convolutional network and LSTM over characters, whereas predictions are given on the word-level.
\newcite{Cao2016} proposed a method for learning both word embeddings and morphological segmentation with a bidirectional recurrent network over characters.
There is also research on performing parsing \cite{Ballesteros2015} and text classification \cite{Zhang2015a} with character-level neural models.
\newcite{Ling2015a} proposed a neural architecture that replaces word embeddings with dynamically-constructed character-based representations. We applied a similar method for operating over characters, but combined them with word embeddings instead of replacing them, as this allows the model to benefit from both approaches.
\newcite{Lample2016} described a model where the character-level representation is combined with word embeddings through concatenation. 
In this work, we proposed an alternative architecture, where the representations are combined using an attention mechanism, and evaluated both approaches on a range of tasks and datasets.
Recently, \newcite{Miyamoto2016} have also described a related method for the task of language modelling, combining characters and word embeddings using gating.

\begin{table*}
\setlength\tabcolsep{9pt}
\begin{tabular}{r|cc|cc|cc} \toprule
 & \multicolumn{2}{c|}{CoNLL03} & \multicolumn{2}{c|}{FCEPUBLIC} & \multicolumn{2}{c}{CHEMDNER} \\ 
 & \# total & \# noemb & \# total & \# noemb & \# total & \# noemb  \\ \midrule 
Word-based & 4,507,658 & 1,230,158 & 2,972,052 & 1,230,252 & 5,862,878 & 1,070,278 \\
Char concat & 4,987,658 & 1,710,158 & 3,452,052 & 1,710,252 & 6,182,878 & 1,390,278 \\
Char attention & 4,687,958 & 1,410,458 & 3,152,352 & 1,410,552 & 5,943,078 & 1,150,478 \\ \bottomrule
\end{tabular}
\caption{Comparison of trainable parameters in each of the neural model architectures. \textit{\# total} shows the total number of parameters; \textit{\# noemb} shows the parameter count excluding word embeddings, as only a small fraction of the embeddings are utilised at every iteration.}
\label{tab:paramsizes}
\end{table*}

\section{Conclusion}

Developments in neural network research allow for model architectures that work well on a wide range of sequence labeling datasets without requiring hand-crafted data.
While word-level representation learning is a powerful tool for automatically discovering useful features, these models still come with certain weaknesses -- rare words have low-quality representations, previously unseen words cannot be modeled at all, and morpheme-level information is not shared with the whole vocabulary.

In this paper, we investigated character-level model components for a sequence labeling architecture, which allow the system to learn useful patterns from sub-word units.
In addition to a bidirectional LSTM operating over words, a separate bidirectional LSTM is used to construct word representations from individual characters. We proposed a novel architecture for combining the character-based representation with the word embedding by using an attention mechanism, allowing the model to dynamically choose which information to use from each information source. In addition, the character-level composition function is augmented with a novel training objective, optimising it to predict representations that are similar to the word embeddings in the model.

The evaluation was performed on 8 different sequence labeling datasets, covering a range of tasks and domains. We found that incorporating character-level information into the model improved performance on every benchmark, indicating that capturing features regarding characters and morphmes is indeed useful in a general-purpose tagging system. In addition, the attention-based model for combining character representations outperformed the concatenation method used in previous work in all evaluations. Even though the proposed method requires fewer parameters, the added ability of controlling how much character-level information is used for each word has led to  improved performance on a range of different tasks.

\bibliographystyle{acl}
\bibliography{charattn}

\begin{thebibliography}{}

\bibitem[\protect\citename{Ballesteros \bgroup et al.\egroup
  }2015]{Ballesteros2015}
Miguel Ballesteros, Chris Dyer, and Noah~A. Smith.
\newblock 2015.
\newblock {Improved Transition-Based Parsing by Modeling Characters instead of
  Words with LSTMs}.
\newblock {\em Proceedings of the 2015 Conference on Empirical Methods in
  Natural Language Processing (EMNLP)}.

\bibitem[\protect\citename{Bergstra \bgroup et al.\egroup }2010]{Bergstra2010}
James Bergstra, Olivier Breuleux, Frederic~Fr{\'{e}}d{\'{e}}ric Bastien, Pascal
  Lamblin, Razvan Pascanu, Guillaume Desjardins, Joseph Turian, David
  Warde-Farley, and Yoshua Bengio.
\newblock 2010.
\newblock {Theano: a CPU and GPU math compiler in Python}.
\newblock {\em Proceedings of the Python for Scientific Computing Conference
  (SciPy)}.

\bibitem[\protect\citename{Campos \bgroup et al.\egroup }2015]{Campos2015}
David Campos, Sergio Matos, and Jose~L. Oliveira.
\newblock 2015.
\newblock {A document processing pipeline for annotating chemical entities in
  scientific documents}.
\newblock {\em Journal of Cheminformatics}, 7.

\bibitem[\protect\citename{Cao and Rei}2016]{Cao2016}
Kris Cao and Marek Rei.
\newblock 2016.
\newblock {A Joint Model for Word Embedding and Word Morphology}.
\newblock In {\em Proceedings of the 1st Workshop on Representation Learning
  for NLP (RepL4NLP-2016)}.

\bibitem[\protect\citename{Collobert \bgroup et al.\egroup
  }2011]{Collobert2011}
Ronan Collobert, Jason Weston, L{\'{e}}on Bottou, Michael Karlen, Koray
  Kavukcuoglu, and Pavel Kuksa.
\newblock 2011.
\newblock {Natural Language Processing (Almost) from Scratch}.
\newblock {\em Journal of Machine Learning Research}, 12.

\bibitem[\protect\citename{Hochreiter and Schmidhuber}1997]{Hochreiter1997}
Sepp Hochreiter and J{\"{u}}rgen Schmidhuber.
\newblock 1997.
\newblock {Long Short-term Memory}.
\newblock {\em Neural Computation}, 9.

\bibitem[\protect\citename{Huang \bgroup et al.\egroup }2015]{Huang2015}
Zhiheng Huang, Wei Xu, and Kai Yu.
\newblock 2015.
\newblock {Bidirectional LSTM-CRF Models for Sequence Tagging}.
\newblock {\em arXiv:1508.01991}.

\bibitem[\protect\citename{Irsoy and Cardie}2014]{Irsoy2014a}
Ozan Irsoy and Claire Cardie.
\newblock 2014.
\newblock {Opinion Mining with Deep Recurrent Neural Networks}.
\newblock In {\em Proceedings of the 2014 Conference on Empirical Methods in
  Natural Language Processing (EMNLP)}.

\bibitem[\protect\citename{Kim \bgroup et al.\egroup }2004]{Kim2004}
Jin-Dong Kim, Tomoko Ohta, Yoshimasa Tsuruoka, Yuka Tateisi, and Nigel Collier.
\newblock 2004.
\newblock {Introduction to the Bio-entity Recognition Task at JNLPBA}.
\newblock {\em Proceedings of the International Joint Workshop on Natural
  Language Processing in Biomedicine and Its Applications}.

\bibitem[\protect\citename{Kim \bgroup et al.\egroup }2016]{Kim2016}
Yoon Kim, Yacine Jernite, David Sontag, and Alexander~M. Rush.
\newblock 2016.
\newblock {Character-Aware Neural Language Models}.
\newblock {\em In Proceedings of the 30th AAAI Conference on Artificial
  Intelligence (AAAI’16)}.

\bibitem[\protect\citename{Krallinger \bgroup et al.\egroup
  }2015]{Krallinger2015}
Martin Krallinger, Florian Leitner, Obdulia Rabal, Miguel Vazquez, Julen
  Oyarzabal, and Alfonso Valencia.
\newblock 2015.
\newblock {CHEMDNER: The drugs and chemical names extraction challenge}.
\newblock {\em Journal of Cheminformatics}, 7(Suppl 1).

\bibitem[\protect\citename{Lafferty \bgroup et al.\egroup }2001]{Lafferty2001}
John Lafferty, Andrew McCallum, and Fernando Pereira.
\newblock 2001.
\newblock {Conditional random fields: Probabilistic models for segmenting and
  labeling sequence data}.
\newblock In {\em Proceedings of the 18th International Conference on Machine
  Learning}.

\bibitem[\protect\citename{Lample \bgroup et al.\egroup }2016]{Lample2016}
Guillaume Lample, Miguel Ballesteros, Sandeep Subramanian, Kazuya Kawakami, and
  Chris Dyer.
\newblock 2016.
\newblock {Neural Architectures for Named Entity Recognition}.
\newblock In {\em Proceedings of NAACL-HLT 2016}.

\bibitem[\protect\citename{Ling \bgroup et al.\egroup }2015a]{Ling2015a}
Wang Ling, Tiago Lu{\'{\i}}s, Lu{\'{\i}}s Marujo, Ram{\'{o}}n~Fernandez
  Astudillo, Silvio Amir, Chris Dyer, Alan~W. Black, and Isabel Trancoso.
\newblock 2015a.
\newblock {Finding Function in Form: Compositional Character Models for Open
  Vocabulary Word Representation}.
\newblock {\em Proceedings of the 2015 Conference on Empirical Methods in
  Natural Language Processing}.

\bibitem[\protect\citename{Ling \bgroup et al.\egroup }2015b]{Ling2015b}
Wang Ling, Isabel Trancoso, Chris Dyer, and Alan~W Black.
\newblock 2015b.
\newblock {Character-based Neural Machine Translation}.
\newblock {\em arXiv preprint arXiv:1511.04586}.

\bibitem[\protect\citename{Marcus \bgroup et al.\egroup }1993]{Marcus1993b}
Mitchell~P. Marcus, Beatrice Santorini, and Mary~Ann Marcinkiewicz.
\newblock 1993.
\newblock {Building a large annotated corpus of English: The Penn Treebank.}
\newblock {\em Computational Linguistics}, 19.

\bibitem[\protect\citename{Mikolov \bgroup et al.\egroup }2013]{Mikolov2013a}
Tom{\'{a}}{\v{s}} Mikolov, Greg Corrado, Kai Chen, and Jeffrey Dean.
\newblock 2013.
\newblock {Efficient Estimation of Word Representations in Vector Space}.
\newblock {\em Proceedings of the International Conference on Learning
  Representations (ICLR 2013)}.

\bibitem[\protect\citename{Miyamoto and Cho}2016]{Miyamoto2016}
Yasumasa Miyamoto and Kyunghyun Cho.
\newblock 2016.
\newblock {Gated Word-Character Recurrent Language Model}.
\newblock {\em arXiv preprint arXiv:1606.01700}.

\bibitem[\protect\citename{Ohta \bgroup et al.\egroup }2002]{Ohta2002}
Tomoko Ohta, Yuka Tateisi, and Jin-Dong Kim.
\newblock 2002.
\newblock {The GENIA corpus: An annotated research abstract corpus in molecular
  biology domain}.
\newblock {\em Proceedings of the second international conference on Human
  Language Technology Research}.

\bibitem[\protect\citename{Rei and Yannakoudakis}2016]{Rei2016}
Marek Rei and Helen Yannakoudakis.
\newblock 2016.
\newblock {Compositional Sequence Labeling Models for Error Detection in
  Learner Writing}.
\newblock In {\em Proceedings of the 54th Annual Meeting of the Association for
  Computational Linguistics}.

\bibitem[\protect\citename{Smith \bgroup et al.\egroup }2008]{Smith2008}
Larry Smith, Lorraine~K Tanabe, Rie~Johnson nee Ando, Cheng-Ju Kuo, I-Fang
  Chung, Chun-Nan Hsu, Yu-Shi Lin, Roman Klinger, Christoph~M Friedrich, Kuzman
  Ganchev, Manabu Torii, Hongfang Liu, Barry Haddow, Craig~A Struble, Richard~J
  Povinelli, Andreas Vlachos, William~A Baumgartner, Lawrence Hunter, Bob
  Carpenter, Richard Tzong-Han Tsai, Hong-Jie Dai, Feng Liu, Yifei Chen,
  Chengjie Sun, Sophia Katrenko, Pieter Adriaans, Christian Blaschke, Rafael
  Torres, Mariana Neves, Preslav Nakov, Anna Divoli, Manuel
  Ma{\~{n}}a-L{\'{o}}pez, Jacinto Mata, and W~John Wilbur.
\newblock 2008.
\newblock {Overview of BioCreative II gene mention recognition.}
\newblock {\em Genome biology}, 9 Suppl 2.

\bibitem[\protect\citename{{Tjong Kim Sang} and
  Buchholz}2000]{TjongKimSang2000}
Erik~F. {Tjong Kim Sang} and Sabine Buchholz.
\newblock 2000.
\newblock {Introduction to the CoNLL-2000 shared task: Chunking}.
\newblock {\em Proceedings of the 2nd Workshop on Learning Language in Logic
  and the 4th Conference on Computational Natural Language Learning}, 7.

\bibitem[\protect\citename{{Tjong Kim Sang} and {De
  Meulder}}2003]{TjongKimSang2003}
Erik~F. {Tjong Kim Sang} and Fien {De Meulder}.
\newblock 2003.
\newblock {Introduction to the CoNLL-2003 Shared Task: Language-Independent
  Named Entity Recognition}.
\newblock In {\em Proceedings of the seventh conference on Natural language
  learning at HLT-NAACL 2003}.

\bibitem[\protect\citename{Tsuruoka \bgroup et al.\egroup }2005]{Tsuruoka2005}
Yoshimasa Tsuruoka, Yuka Tateishi, Jin~Dong Kim, Tomoko Ohta, John McNaught,
  Sophia Ananiadou, and Jun'ichi Tsujii.
\newblock 2005.
\newblock {Developing a robust part-of-speech tagger for biomedical text}.
\newblock In {\em Proceedings of Panhellenic Conference on Informatics}.

\bibitem[\protect\citename{Yannakoudakis \bgroup et al.\egroup
  }2011]{Yannakoudakis2011}
Helen Yannakoudakis, Ted Briscoe, and Ben Medlock.
\newblock 2011.
\newblock {A New Dataset and Method for Automatically Grading ESOL Texts}.
\newblock In {\em Proceedings of the 49th Annual Meeting of the Association for
  Computational Linguistics: Human Language Technologies}.

\bibitem[\protect\citename{Zeiler}2012]{Zeiler2012}
Matthew~D. Zeiler.
\newblock 2012.
\newblock {ADADELTA: An Adaptive Learning Rate Method}.
\newblock {\em arXiv preprint arXiv:1212.5701}.

\bibitem[\protect\citename{Zhang \bgroup et al.\egroup }2015]{Zhang2015a}
Xiang Zhang, Junbo Zhao, and Yann LeCun.
\newblock 2015.
\newblock {Character-level Convolutional Networks for Text Classification}.
\newblock In {\em Advances in Neural Information Processing Systems}.

\bibitem[\protect\citename{Zhou and Su}2004]{Zhou2004}
GuoDong Zhou and Jian Su.
\newblock 2004.
\newblock {Exploring Deep Knowledge Resources in Biomedical Name Recognition}.
\newblock {\em Workshop on Natural Language Processing in Biomedicine and Its
  Applications at COLING}.

\end{thebibliography}

\end{document}